\documentclass[runningheads]{llncs}
\usepackage{graphicx,verbatim}
\usepackage{multirow}
\usepackage{makecell}
\usepackage{booktabs}
\usepackage{cite}
\usepackage{amssymb}
\usepackage{amsmath}
\usepackage[T1]{fontenc}

\newcommand{\cmark}{\checkmark}
\newcommand{\xmark}{--}

\begin{document}
\title{Self-Supervised Multiview Xray Matching}
%

\author{Mohamad Dabboussi\inst{1,2} \and Malo Huard \inst{2} \and Yann Gousseau \inst{1} \and Pietro Gori \inst{1} }

\authorrunning{M. Dabboussi et al.}

\institute{ Telecom Paris, Palaiseau, France\\
\and Milvue, Paris, France
}
    
\maketitle              
\begin{abstract}
Accurate interpretation of multi-view radiographs is crucial for diagnosing fractures, muscular injuries, and other anomalies. While significant advances have been made in AI-based analysis of single images, current methods often struggle to establish robust correspondences between different X-ray views, an essential capability for precise clinical evaluations. In this work, we present a novel self-supervised pipeline that eliminates the need for manual annotation by automatically generating a many-to-many correspondence matrix between synthetic X-ray views. This is achieved using digitally reconstructed radiographs (DRR), which are automatically derived from unannotated CT volumes. Our approach incorporates a transformer-based training phase to accurately predict correspondences across two or more X-ray views. Furthermore, we demonstrate that learning correspondences among synthetic X-ray views can be leveraged as a pretraining strategy to enhance automatic multi-view fracture detection on real data. Extensive evaluations on both synthetic and real X-ray datasets show that incorporating correspondences improves performance in multi-view fracture classification.


\begin{figure}
\centering
\includegraphics[width=1.0\textwidth]{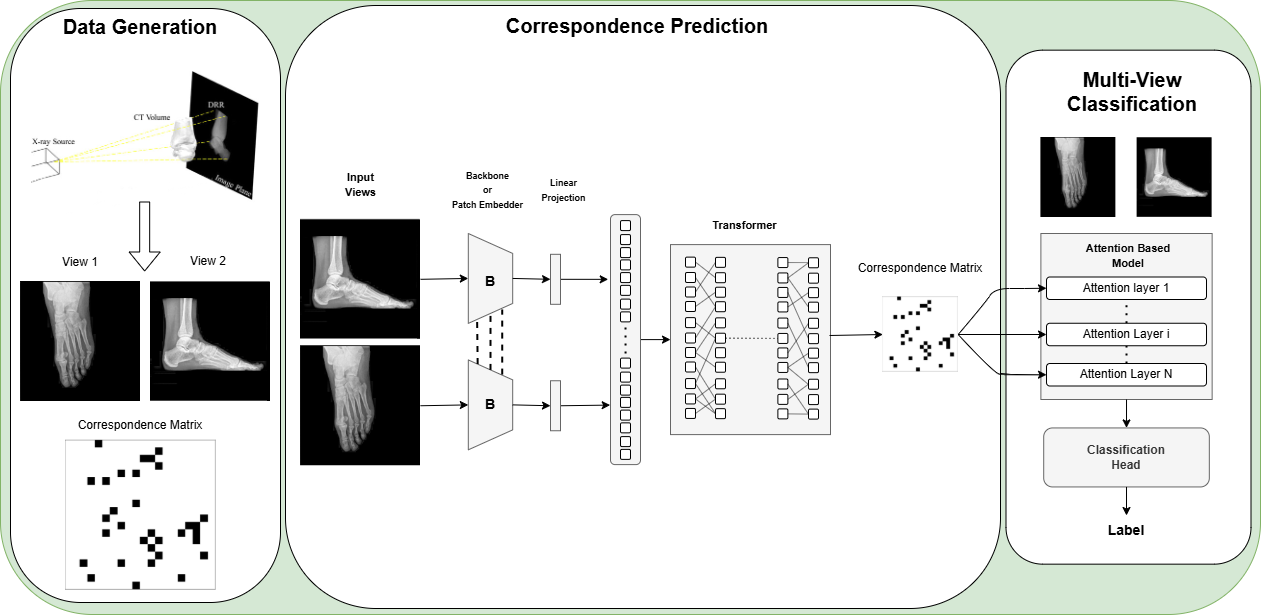}
\caption{End-to-end self-supervised pipeline for predicting X-ray correspondences and multi-view fracture classification. (Left) From a CT volume we generate multiple 2D DRRs (digitally reconstructed radiographs) with a correspondence matrix linking matching points. (Middle) A backbone network extracts features from each view, processed by a Transformer to predict a Correspondence Matrix via attention mechanisms. (Right) The correspondence guides attention for fracture detection.}
\label{fig:Pipeline}
\end{figure}

\end{abstract}

\section{Introduction}
Accurate diagnosis in radiology often relies on the complementary information provided by multiple X-ray views that are acquired from various angles to ensure a comprehensive evaluation. 
In practice, radiologists examine all these views to confirm the presence and extent of lesions, thus increasing diagnostic confidence.

\noindent \textbf{Motivation }
Multi-view imaging is essential in radiology, as each X-ray projection provides unique information that aids in detecting subtle abnormalities and improving overall evaluation. However, the process is time-consuming, requires expertise, and is prone to errors. These challenges highlight the need for automated methods capable of effectively handling multi-view data to support clinical decision-making.

\begin{figure}
\centering
\includegraphics[width=.6\textwidth]{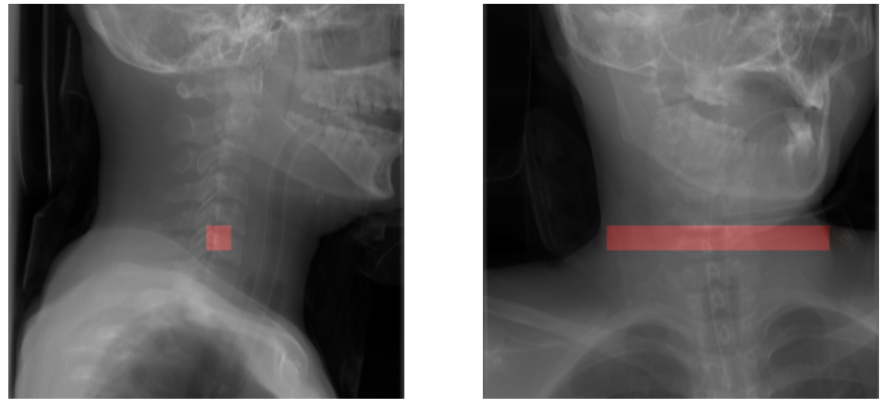}
\caption{Many-to-many matching between two X-ray views of the same subject. The red patch in the left view corresponds to the red bar in the right view, illustrating multiple matching pixels across views.}
\label{fig:Transparent_Matching}
\end{figure}

\noindent \textbf{Problem Statement }
While deep learning achieved impressive results  in single-view medical imaging, extending it to multi-view X-ray interpretation remains challenging. Progress is hindered by two key issues: (1) the scarcity of annotated multi-view datasets, especially beyond chest X-rays, and (2) the complexity of establishing correspondences between views. Traditional image matching techniques focus on one-to-one mappings in natural images, whereas X-rays involve complex, many-to-many relationships due to the cumulative nature of pixel intensities along the X-ray path  (see Fig.~\ref{fig:Transparent_Matching}). This makes it challenging to determine whether abnormalities across views correspond to the same pathology. There is a need for an automated method to estimate accurate multi-view correspondences without extensive manual annotation, ultimately aiding radiologists and improving lesion/fracture detection.

To this end, our work introduces the following \textbf{contributions}:\\
 - \textbf{DRR-based Correspondence Generation:} We generate paired simulated X-rays views and patch-level correspondence matrices from unannotated CT volumes, thereby eliminating the need for large annotated X-ray datasets.\\
 - \textbf{Self-supervised Pre-training:} We leverage the correspondence prediction task as a self-supervised method to enhance multi-view feature learning.\\
- \textbf{Transformer Integration:} We incorporate correspondence information within transformer-based architectures to improve fracture classification performance.

\section{Related Works}

\textbf{DRR-based Dataset Generation }
Digitally Reconstructed Radiographs (DRR) synthesize X-ray images from CT volumes. Early ray-tracing methods ~\cite{siddon1985prism, joseph1982improved} optimized efficiency, while recent deep learning approaches ~\cite{unberath2018deepdrr, dhont2020realdrr} improved realism. DRRs have been used for: (1) 3D CT reconstruction from limited X-ray views ~\cite{corona2022mednerf}, (2) 2D/3D image registration ~\cite{Zhang1023registration, shrestha2023x, wu2023multi}, and (3) generating labeled datasets for segmentation using manually labeled 3D CT volumes ~\cite{killeen2024fluorosam, zhang2018task, kasten2020end}. Unlike prior work, we leverage DRR to automatically generate correspondence annotations from \textit{unannotated} CT volumes, enabling large-scale, annotated (i.e., correspondences), multi-view X-ray datasets across diverse anatomical regions.

\noindent \textbf{Image Matching and Correspondence Estimation }
Traditional feature-based image matching \cite{lowe2004distinctive, herbert2008surf} has been largely replaced by deep learning approaches. Sparse keypoint methods \cite{Sarlin_2020_CVPR, Lindenberger_2023_ICCV} and dense matching techniques \cite{Sun_2021_CVPR, edstedt2024roma, chen2022aspanformer} perform well in natural image tasks but assume one-to-one pairwise matching, which suffices for tasks involving two views with homographies, such as camera pose estimation. Multi-view X-ray imaging, however, requires many-to-many mappings due to anatomical transparency and overlapping structures. Existing methods struggle in this context, highlighting the need for specialized solutions in X-ray correspondence estimation.

To address these challenges, we propose a self-supervised pipeline that generates from unannotated CT volumes a large, diverse dataset of synthetic X-rays with automatically derived correspondence matrices, eliminating the need for manual annotations. Unlike traditional pairwise matching methods, our approach learns many-to-many correspondences across multiple views, enabling a more comprehensive understanding of their complex spatial relationships and advancing automated multi-view X-ray analysis.

\section{Method}
\label{section_method}
We introduce a novel approach for generating correspondence ground truth using Digitally Reconstructed Radiographs (DRR) from unannotated CT scans. Synthetic X-ray views are first created along with their corresponding correspondence matrices. These are then used as supervision for training a deep-learning model, which takes two synthetic X-ray views as input and predicts their correspondence matrix. 

To evaluate  the effectiveness of our approach, we conduct experiments on both synthetic and real multi-view X-ray datasets, demonstrating the effectiveness of correspondence learning not only as a pretraining strategy but also as a mechanism for attention guidance in multi-view fracture classification.

\subsection{Correspondence Ground Truth Generation}
\begin{figure} 
\centering
\includegraphics[width=1.0\textwidth]{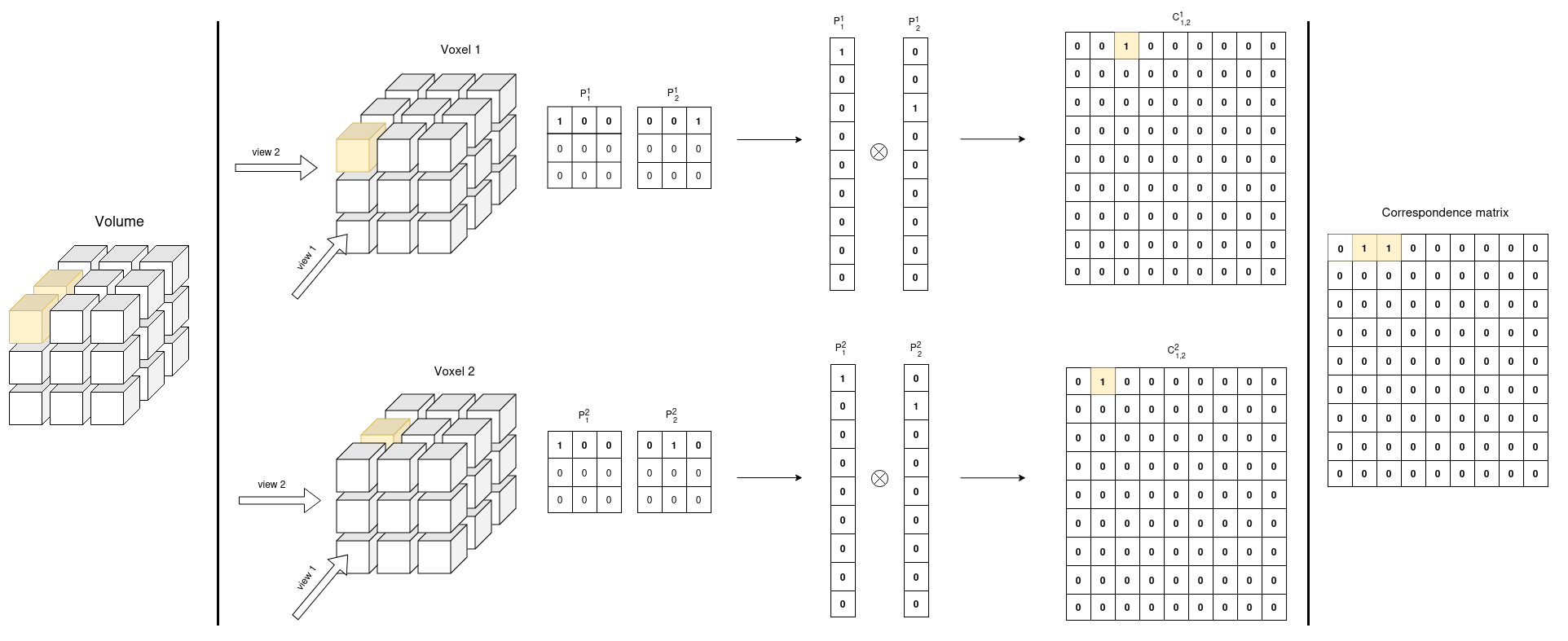}
\caption{Generating a correspondence matrix for a $3\times3\times3$ volume with a highlighted $1\times1\times2$ cube (yellow) using two orthogonal views. Each row shows the projection of a yellow voxel \( v \) onto both views. The resulting projection matrices, $P^v_1$ and $P^v_2$, are flattened, and their outer product forms $C^v_{1,2}$. The final correspondence matrix is obtained by taking the maximum over all $C^v_{1,2}$ matrices.}
\label{fig:corr_3d_90}
\end{figure}

Let \( V \in \mathbb{R}^{H \times W \times L} \) denote a CT volume consisting of \( N \) voxels. For each non-air voxel \( v \) at position \( (x,y,z) \), obtained by thresholding the CT volume, we project it onto two distinct view directions using the Joseph method \cite{joseph1982improved} to generate DRRs. This results in two projection matrices, \( P_1^v, P_2^v \in \mathbb{R}^{H \times W} \), which encode the accumulated intensity values along rays traced from the X-ray source to the detector plane (see Figure~\ref{fig:corr_3d_90} for a visual explanation). We then define their flattened representations as: $
\mathbf{p}_1^v = \mathit{flat}(P_1^v) \in \mathbb{R}^{HW}$ and $\mathbf{p}_2^v = \mathit{flat}(P_2^v) \in \mathbb{R}^{HW}$, 
where \(\mathit{flat}(\cdot)\) denotes the flattening of a matrix into a vector. 
The voxel-specific correspondence matrix is then given by the outer product: $
\mathbf{C}_{1,2}^v = \mathbf{p}_1^v \mathbf{p}_2^{v\top} \in \mathbb{R}^{HW \times HW}
$. 
The final correspondence matrix is obtained by taking the element-wise maximum over all voxels:  $
\mathbf{C}_{1,2} = \max_{v} \left( \mathbf{C}_{1,2}^v \right)$.

To address computational challenges posed by high-resolution CT volumes (e.g., \(256 \times 256 \times 256\)), 
 we perform patch-level (and not pixel-level) correspondence estimation. Specifically, we downsample the volume by a factor of \( k=16 \), thereby computing the correspondence matrix at a coarser, patch-wise resolution.

\begin{figure}
\centering
\includegraphics[width=.3\textwidth]{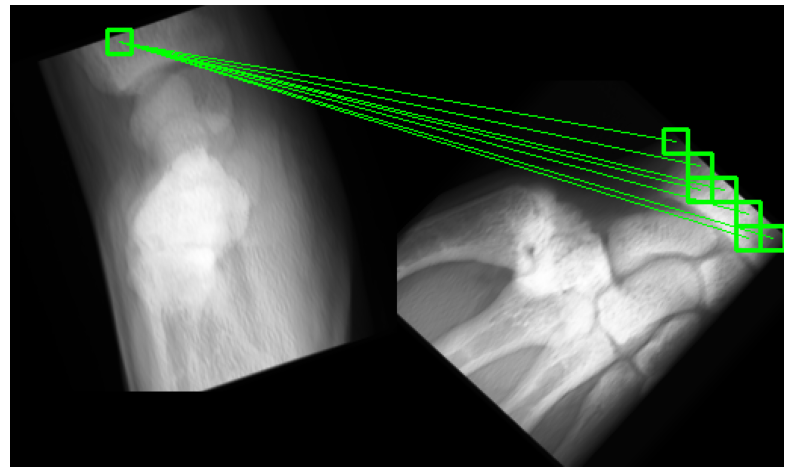}
\includegraphics[width=.3\textwidth]{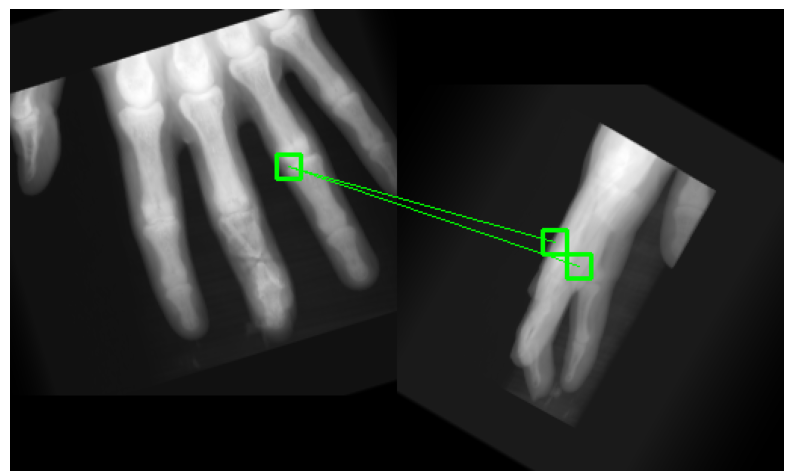}
\includegraphics[width=.3\textwidth]{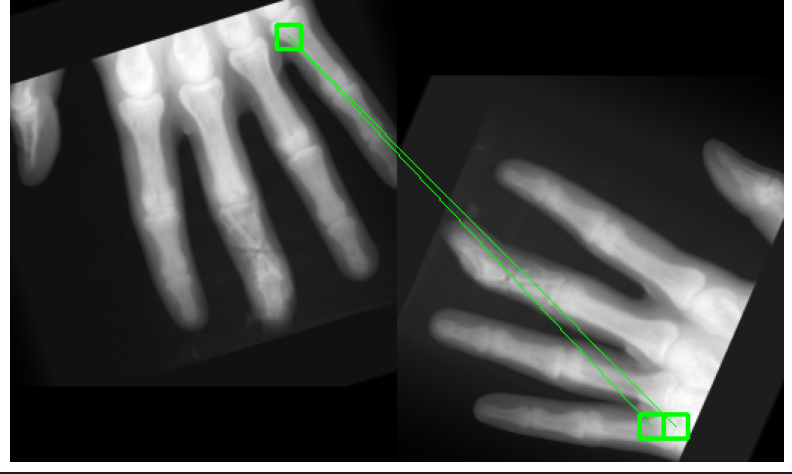}
\caption{Visualization of generated hand correspondences annotations.}
\label{fig:Viz_corr}
\end{figure}

\subsection{Correspondence Prediction}
We formulate correspondence prediction as a similarity assessment between features extracted from patches across different views. A pretrained backbone network, denoted as \( f \), extracts feature maps from each image, which are then projected into a lower-dimensional embedding space. The spatial grid of these feature maps is flattened into a sequence of patch embeddings, where each embedding represents a distinct image region. Afterwards, patch embeddings from multiple views are concatenated along the sequence dimension and processed by a transformer module. This module employs self-attention to capture both intra- and inter-view interactions, effectively encoding correspondence information. Correspondences are then determined by computing a normalized dot product between feature pairs, forming a correspondence matrix. This approach naturally extends to multi-view scenarios with more than two images.

\subsection{Pre-training through correspondence prediction}
As we will show in Section~\ref{sec:experiments}, we can use correspondence prediction as a self-supervised pretraining step for an auxiliary multi-view X-ray classification task. This approach encourages the model to learn robust features that capture shared anatomical information while leveraging the redundancy in multi-view data.

\subsection{Attention Guidance  in Multi-View X-ray classification}

 To leverage cross-view correspondence information in our multi-view X-ray classification framework, we can also integrate the correspondence matrix directly into the transformer attention mechanism. Specifically, the correspondence matrix is employed as an attention bias, which guide the model to focus on corresponding patches across views during the classification task.
 
 Let $Q$ and $K$ denote the query and key matrices, respectively, and let $d$ be the dimensionality of each attention head. The standard scaled dot-product attention is computed as $A = \frac{QK^\top}{\sqrt{d}}$ (Eq. 1). Given a correspondence matrix $C$, the attention scores are adjusted as $A' = A + \alpha\, C$ (Eq. 2), where $\alpha$ is a learnable scaling parameter. Finally, the modified scores $A'$ are normalized via the softmax function to yield the attention probabilities.

\section{Experiments}
\label{sec:experiments}
\subsection{Datasets and Implementation Details}
\textbf{Train Simulated Dataset } We generated correspondence matrices using our proposed method. In particular, 207,600 data samples were generated from 175 CT volumes by varying projection angles, the distance between the volume and the source, as well as by extracting appropriate crops from the CT volumes.

\noindent \textbf{Test Simulated X-ray Correspondence Dataset }
From 42 CT volumes of extremity, not seen during training, 16,920 pairs of views along with their corresponding correspondence matrices were generated using the same variation parameters as during training.

\noindent \textbf{Real X-ray Correspondence Dataset }
This dataset comprises 347 samples, each with two views. We applied a 60\%/10\%/30\% split for training, validation, and testing. Each sample includes only three annotated positive correspondences, along with 100 negative correspondences.

\noindent \textbf{MURA Public Dataset } The MURA dataset  ~\cite{rajpurkar2017mura} comprises studies from various anatomical regions. For our purposes, we selected studies of the elbow, forearm, hand, and wrist, which correspond to the regions on which the correspondence model was trained. The final dataset includes 5,511 studies. Each study has multiple views, we used 2 views in our experiments.

\noindent \textbf{Private Dataset } This dataset mainly contains multiview X-rays of the hand, forearm, foot, and knee. It has a total of 5,653 studies. Each study has multiple views, we used 2 views in our experiments.

\noindent \textbf{Implementation Details }
Our framework processes two input views, each of size 256×256 pixels. We use a pre-trained ResNet-50 \cite{he2016deep} backbone for feature extraction, followed by a transformer with Rotary Positional Encoding (RoPE \cite{su2024roformer}) to capture spatial relationships between patches. Training utilizes the Adam optimizer with a cosine annealing learning rate scheduler. The initial learning rate is set to $1\times10^{-4}$ for pre-training on simulated correspondence data and reduced to $1\times10^{-5}$ when fine-tuning on partially annotated real data. The model is trained with a batch size of 16. For loss, we use mean squared error (MSE) for the correspondence task and binary cross-entropy for classification. Data augmentation techniques, including random adjustments to brightness, contrast, and color inversion, are applied to enhance model robustness.

\subsection{Results and Discussion}

\begin{table}[htbp]
  \centering
    \caption{Model Performance on Correspondence Simulated Test Dataset.}
  \begin{tabular}{c c c c c c}
    \toprule
    \textbf{Attention Model} & \textbf{Message Pass} & 
    \textbf{MSE} & \textbf{Precision} & \textbf{Recall} & \textbf{AP} \\
    \midrule
    \xmark   & \xmark   & 2.25 $\times$ 10$^{-3}$ &  49.1 & 68.0 & 54.7 \\
    Superglue Module & \cmark & 1.33 $\times$ 10$^{-3}$ & 70.3 & 75.6 & 80.1 \\
    LoFTR Module & \cmark & 1.24 $\times$ 10$^{-3}$ & 74.2 & 78.1 & 83.5 \\
    Standard Transformer & \cmark   & 1.19 $\times$ 10$^{-3}$ & 75.5 & 79.4 & 84.1 \\
    Standard Transformer & \xmark   & \textbf{1.09 $\times$ 10$^{-3}$}  & \textbf{77.0} & \textbf{81.5} & \textbf{85.0}  \\
    \bottomrule
  \end{tabular}
  \label{tab:table_simulated_test}
\end{table}

\noindent \textbf{Correspondence prediction - Simulated data } Table \ref{tab:table_simulated_test} summarizes the performance of various models for our patch-level correspondence prediction task on a simulated test dataset. We use ResNet-50 as the backbone for feature extraction and compute the correspondence matrix using normalized dot product.
Our experiments compare different attention modules: graph-based message passing methods (e.g., SuperGlue \cite{Sarlin_2020_CVPR} and LoFTR \cite{Sun_2021_CVPR}) and a standard transformer attention module. The evaluation metrics include mean squared error (MSE), precision, recall, and average precision (AP).
Notably, our approach, the Standard Transformer \cite{vaswani2017attention} without message passing, achieved the best performance across all metrics, with the lowest MSE and the highest AP. This suggests that the transformer’s flexible attention mechanism enables more effective inter-patch communication compared to the more constrained graph-based approaches. Since our goal is to predict correspondences at the patch level, acknowledging that abnormalities typically span groups of pixels rather than isolated pixels, it is natural to focus on patches. In contrast, a CNN-only baseline without any attention mechanism performed considerably worse, highlighting the importance of attention mechanisms in learning robust patch-level correspondences.

\begin{table*}
    \centering
     \caption{Model Performance on the Correspondence Test Dataset of Real X-ray.}
    \begin{tabular}{c c c c c c}
        \toprule
        \textbf{Method} & \textbf{Backbone} & \textbf{Model} & \textbf{Precision} & \textbf{Recall} & \textbf{AP} \\
        \midrule
        \multirow{2}{*}{Zero Shot} 
            & Res50  & LoFTR module & 32.5 &  2.7 & 9.4 \\
            & Res50  & LoFTR$^{*}$ module & 28.3 &  26.3 & 15.9 \\
            & DinoV2-G & \xmark & 3.0   & 15.2  & 1.7 \\
        \midrule
        \multirow{2}{*}{Fine-tuning}
            & Res50  & LoFTR$^{*}$ module & 36.8 &  18.0 & 16.1 \\
            & DinoV2-G & Multi Layer Perceptron & 55.23   & 51.6  & 42.2 \\
        \midrule
        \multirow{1}{*}{Pre-train + Fine-tuning}
            & Res50  & Standard Transformer & \textbf{72.3}   & \textbf{87.1}  & \textbf{83.8} \\
        \bottomrule
    \end{tabular}
    \label{tab:table_real_test}
\end{table*}

\noindent \textbf{Correspondence prediction - Real data }Table \ref{tab:table_real_test} evaluates our method for correspondence matching on a real X-ray dataset with partial annotations, comparing it against existing approaches in both zero-shot and fine-tuned settings. The goal is to establish reliable point correspondences with minimal annotation. In the zero-shot setting, the pre-trained LoFTR model \cite{Sun_2021_CVPR} with a ResNet50 backbone struggles due to domain shift and its one-to-one matching strategy, resulting in low recall (2.7) and modest AP (9.4). We introduce a variant LoFTR$^*$, which incorporates a normalized dot-product and thresholding correspondence head for multi-to-multi matching, slightly improving recall (26.3) and AP (15.9), though the domain gap remains significant. Fine-tuning LoFTR$^*$ improves performance (AP: 16.1), but a stronger baseline is obtained by freezing a 1.1B-parameter DinoV2-G model \cite{oquab2023dinov2} and fine-tuning an MLP head, achieving much better results (AP: 42.2). Our approach, a transformer-based model pre-trained and fine-tuned with only 24M parameters, outperforms all other methods, achieving the highest AP (83.8).

\begin{table*}[ht]
    \centering
        \caption{Classification Performance on the MURA and the Private Dataset.}
    \begin{tabular}{c c c c c c c c c}
        \toprule
     & \makecell{\textbf{Correspondence}\\\textbf{Pretraining}} & \makecell{\textbf{Attention}\\\textbf{Guidance}} & \textbf{Fusion} & \textbf{Accuracy} & \textbf{Precision} & \textbf{Recall} & \textbf{Kappa} \\
        \midrule
        \multirow{6}{*}{\rotatebox{90}{{\centering MURA}}}
            & \xmark  & \xmark & Single & $73.3\pm0.3$   & $75.4\pm0.2$  & $65.4\pm0.2$ & $0.45$ \\
            & \xmark  & \xmark & Late & $78.7\pm0.2$   & $83.3\pm0.1$  & $67.4\pm0.2$ & $0.56$ \\
            & \xmark  & \xmark & Early & $75.8\pm0.2$   & $78.5\pm0.1$  & $65.4\pm0.2$ & $0.49$ \\
            & \cmark & \xmark & Early & $\underline{80.1\pm0.1}$ & $\underline{83.4\pm0.1}$  & $\underline{73.6\pm0.2}$ & $\underline{0.58}$ \\
            & \cmark & \cmark & Early & $\textbf{80.6}\pm\textbf{0.1}$ & $\textbf{84.8}\pm\textbf{0.1}$ & $\textbf{74.4}\pm\textbf{0.2}$ & $\textbf{0.59}$ \\
        \midrule
        \multirow{6}{*}{\rotatebox{90}{{\centering Private}}}
            & \xmark  & \xmark & Single & $68.8\pm0.2$   & $53.6\pm0.2$  & $41.7\pm0.2$ & $0.30$ \\
            & \xmark  & \xmark & Late & $74.2\pm0.2$   & $59.5\pm0.2$  & $\underline{50.1\pm0.2}$ & $0.36$ \\
            & \xmark  & \xmark & Early &  $71.1\pm0.1$ & $54.4\pm0.1$ & $44.2\pm0.1$ & $0.32$ \\
            & \cmark & \xmark & Early & $\underline{75.0\pm0.1}$ & $\underline{59.7\pm0.1}$ & $49.2\pm0.2$ & $\underline{0.37}$ \\
            & \cmark & \cmark & Early & $\textbf{76.2}\pm\textbf{0.1}$ & $\textbf{59.9}\pm\textbf{0.1}$ & $\textbf{52.6}\pm\textbf{0.2}$ & $\textbf{0.39}$ \\
        \bottomrule
    \end{tabular}
    \label{tab:classification}
\end{table*}

\noindent \textbf{Multi-View X-ray classification }
Table \ref{tab:classification} evaluates the impact of correspondence pretraining and attention guidance on multi-view X-ray classification. This experiment demonstrates the benefits of pretraining a model on a correspondence task before applying it to classification, as well as the advantages of using the correspondence mask to guide attention. We used a transformer-based model (e.g., ViT-S \cite{dosovitskiy2020image}) trained on both public and private datasets. We compare early fusion, where patch embeddings from multiple views are concatenated before input to the transformer, and late fusion, where scores are aggregated after independent processing. Multi-view fusion significantly outperforms single-view methods, with early fusion achieving higher accuracy than late fusion.

Pretraining on correspondence further boosts performance, despite all models being initialized with ImageNet weights. Notably, early fusion with pretraining improves accuracy from 75.8\% to 80.1\% on the public dataset. Additionally, incorporating the correspondence matrix as an attention bias further enhances results, yielding the highest accuracy (80.6\% public, 76.2\% private).

\section{Conclusion}
 We introduced a new framework for multi-view X-ray analysis that leverages self-supervised correspondence learning to improve both correspondence matching on real X-ray images and multi-view fracture detection. Our experiments show that pretraining on correspondence significantly enhances classification accuracy, especially when integrating correspondence information into transformer-based architectures. As demonstrated in our fracture classification task, the no-annotation correspondence method we proposed opens up numerous use cases in multi-view X-ray tasks. This work provides a novel approach that can be applied to a variety of X-ray tasks, advancing the field of multi-view medical image analysis.

\bibliographystyle{splncs04}
\bibliography{references}

\end{document}